\documentclass{ecai} 

\usepackage{latexsym}
\usepackage{amssymb}
\usepackage{amsmath}
\usepackage{amsthm}
\usepackage{booktabs}
\usepackage{enumitem}
\usepackage{graphicx}
\usepackage{color}
\usepackage{algorithm} 
\usepackage{algorithmic} 
\usepackage{xcolor} 
\usepackage{tabularx}
\usepackage{makecell}
\usepackage{multirow}
\usepackage{ragged2e}
\usepackage{bm}
\usepackage{url}

\clearpage

\newcommand{\BibTeX}{B\kern-.05em{\sc i\kern-.025em b}\kern-.08em\TeX}

\definecolor{darkgreen}{RGB}{0, 150, 0}

\begin{document}


\begin{frontmatter}


\paperid{355} 


\title{Video2Reward: Generating Reward Function from Videos for Legged Robot Behavior Learning}




\author[A]{\fnms{Runhao}~\snm{Zeng}}
\author[B]{\fnms{Dingjie}~\snm{Zhou}}
\author[B]{\fnms{Qiwei}~\snm{Liang}}
\author[C]{\fnms{Junlin}~\snm{Liu}}
\author[B]{\fnms{Hui}~\snm{Li}}
\author[D]{\fnms{Changxin}~\snm{Huang}\thanks{Corresponding Author. Email: huangchx@szu.edu.cn}}
\author[D]{\fnms{Jianqiang}~\snm{Li}} 
\author[A]{\fnms{Xiping}~\snm{Hu}} 
\author[E]{\fnms{Fuchun}~\snm{Sun}} 

\address[A]{Artificial Intelligence Research Institute, Shenzhen MSU-BIT University, China}
\address[B]{College of Mechatronics and Control Engineering, Shenzhen University}
\address[C]{College of Computer Science and Software Engineering, Shenzhen University}
\address[D]{National Engineering Laboratory for Big Data System Computing Technology, Shenzhen University}
\address[E]{Department of Computer Science and Technology, Tsinghua University}


\begin{abstract}
Learning behavior in legged robots presents a significant challenge due to its inherent instability and complex constraints. Recent research has proposed the use of a large language model (LLM) to generate reward functions in reinforcement learning, thereby replacing the need for manually designed rewards by experts. However, this approach, which relies on textual descriptions to define learning objectives, fails to achieve controllable and precise behavior learning with clear directionality. In this paper, we introduce a new \textbf{video2reward} method, which directly generates reward functions from videos depicting the behaviors to be mimicked and learned. Specifically, we first process videos containing the target behaviors, converting the motion information of individuals in the videos into keypoint trajectories represented as coordinates through a video2text transforming module. These trajectories are then fed into an LLM to generate the reward function, which in turn is used to train the policy. To enhance the quality of the reward function, we develop a video-assisted iterative reward refinement scheme that visually assesses the learned behaviors and provides textual feedback to the LLM. This feedback guides the LLM to continually refine the reward function, ultimately facilitating more efficient behavior learning. Experimental results on tasks involving bipedal and quadrupedal robot motion control demonstrate that our method surpasses the performance of state-of-the-art LLM-based reward generation methods by over 37.6\% in terms of human normalized score. More importantly, by switching video inputs, we find our method can rapidly learn diverse motion behaviors such as walking and running.

\end{abstract}

\end{frontmatter}


\section{Introduction}

Robot behavior learning explores how robots autonomously acquire and improve their behaviors through interactions with their environment, enabling them to adapt to a variety of tasks and scenarios \cite{schoner1995dynamics, beer1995dynamical}. Existing research on robot behavior learning has found widespread application in numerous robotic tasks, such as manipulations \cite{kroemer2021review, rozo2013robot} and autonomous driving \cite{paniego2023model, graves2020learning}. However, due to their complex dynamic mechanisms, instability, and multiple constraints, legged robots present significant challenges in behavior learning.

Current research predominantly focuses on two methodologies: imitation learning and reinforcement learning. Imitation learning involves learning behavior strategies by observing and imitating the demonstrations of experts or other intelligent agents \cite{hussein2017imitation}. This method often employs deep neural networks to learn the associations between inputs and control signals within extensive demonstration datasets \cite{xie2020learning,imi1,dart,siekmann2020learning}. 
For instance, Bohez et al. achieve natural and agile gaits by imitating the motion trajectories of animals or humans \cite{bohez2022imitate}, and Koch et al. improve control accuracy and stability by imitating the output of a high-level controller or optimizer \cite{koch2019reinforcement}.
However, the performance of imitation learning is limited by the quality, quantity, and diversity of the demonstration data. For example, the demonstration data may not cover certain corner cases, potentially leading to catastrophic failures in robots under such circumstances. Recently, reinforcement learning (RL) has made significant strides in addressing complex behavior learning challenges in legged robots \cite{siekmann2020learning, xie2020learning, aractingi2023controlling}. The core of RL involves training robots to take actions that maximize expected cumulative rewards. However, due to the instability and multiple constraints of legged robots, their reward functions require meticulous design. Typically, reward functions consist of multiple components \cite{zhang2019teach}, each describing an aspect of performance in the walking task, such as maintaining body balance, preserving periodic gait, reducing motor wear, etc. The design of these rewards heavily relies on human experience and needs to be redesigned when changes occur in the robot's structure or dynamics, which is undoubtedly time-consuming and labor-intensive.

To mitigate this issue, some researchers have proposed using the zero-shot generative capabilities of large language model (LLM) to automatically design robot reward functions~\cite{ma2023eureka, xie2023text2reward, yu2023language}. For instance, Ma et al.~\cite{ma2023eureka} introduced the Eureka framework, which generates reward function code by providing descriptions of robotic tasks and environments to LLM. Reinforcement learning strategies trained with such rewards successfully control robots to accomplish specific tasks. However, using natural language alone makes it difficult to express desired robotic behaviors with fine granularity/precision. For example, employing LLM to design rewards for a quadrupedal robot’s walking might involve descriptions like ``we expect the robotic dog to pace with natural posture", where ``pacing" is abstract and fails to accurately capture specific motion trajectories. Consequently, the abstract nature of these language descriptions leads to difficult-to-control outcomes in robotic behavior learning. Moreover, relying solely on language descriptions can result in robotic actions that are rigid and unnatural, producing movements that natural animals would not perform, such as a robot executing small, hop-like steps with both legs extended, which does not conform to the typical motion of natural dogs.

\begin{figure}[t]
\centering
\includegraphics[width=\columnwidth]{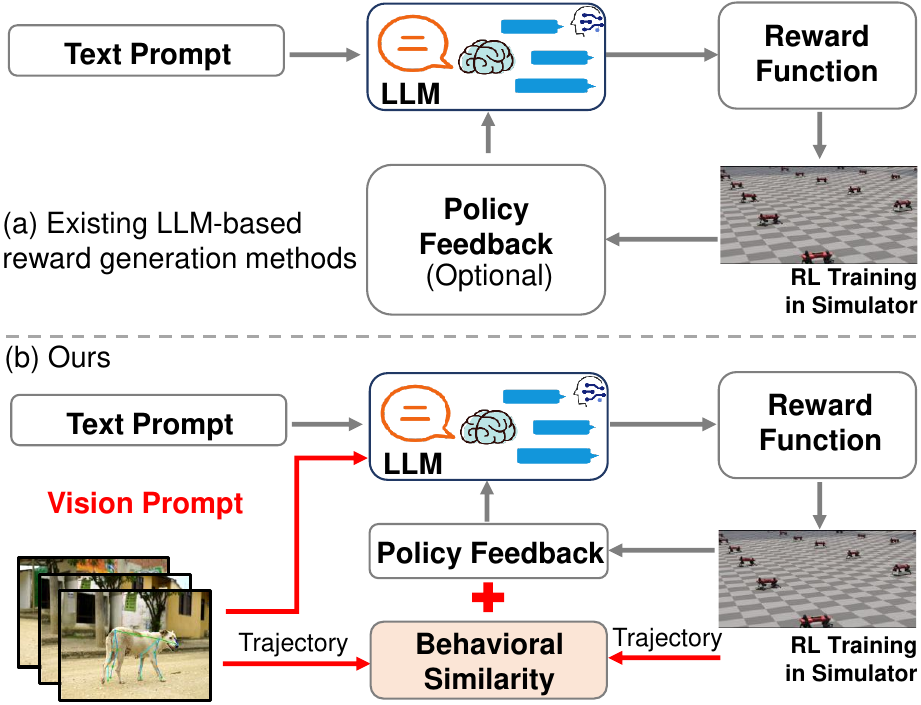}
\vspace{-0.5cm}
\caption{Existing large language model-based reward generation methods~\cite{ma2023eureka,yu2023language} rely solely on text as input, which is inefficient due to the difficulty of adequately describing complex behaviors. Our method directly uses videos with target behaviors as inputs, which can be also used to visually assess the learning outcomes of the agent's behavior, providing feedback to the LLM to facilitate efficient reward generation.}
\vspace{0.8cm}
\label{fig:fig1}
\end{figure}

In this paper, we propose a new method that uses LLM to automatically generate reward functions based on video input. We aim to address two pivotal questions: 1) how can behaviors depicted in videos be processed into a format that LLM can use to generate rewards? 2) how can videos be used to evaluate the effectiveness of reward functions and provide feedback for their refinement? \textbf{First}, we introduce a video-to-text transformation module that converts the motion characteristics of animals or humans into keypoint trajectories. Then these trajectories are translated into textual descriptions suitable for LLM input, ensuring that the resulting reward functions closely correspond to specific behaviors. \textbf{Second}, we propose a video-assisted iterative reward function refinement scheme. This involves visually comparing the learned behaviors with the target behaviors displayed in the videos to evaluate the reward functions, creating feedback that is re-input into the LLM to refine the reward functions. Our method, shown in Figure \ref{fig:fig1}, demonstrates significant improvements over approaches that rely solely on textual descriptions, through experiments on bipedal and quadrupedal robots learning various behaviors, including ambling and running. Moreover, it is capable of adapting to different behaviors based on the input videos. We believe that this research offers new insights and tools for behavior learning, enriching the field with a more dynamic and interactive approach to developing reward functions that are deeply integrated with visual data.


%
The contributions of this work are summarized as follows:
\begin{itemize}
\item \textbf{A More Controllable Approach for Robot Behavior Learning:} We propose a new method that uses video to guide the generation of robot reward functions through LLM. By extracting keypoints motion trajectories from videos, this framework guides the LLM to generate executable reward functions that are consistent with the movement characteristics of objects depicted in the videos. This framework allows for control over the robot's learning of various behaviors by altering the input video.
\item \textbf{New Video-Assisted Feedback Scheme for Reward Generation:} By incorporating videos as learning targets, we propose a robot behavior learning evaluation mechanism based on video feedback. This approach allows for a more visually intuitive assessment of the discrepancies between the behaviors enacted by reinforcement learning policies and the desired behaviors. Consequently, it guides the generation of more appropriate rewards by LLM, resulting in robot behaviors that are more natural and fluid, and that comply with physical laws.
\item \textbf{Improved Performance in Legged Robot Behavior Learning Tasks Compared to Existing LLM-based Reward Generation Methods:} Experiments conducted in the IsaacGym~\cite{makoviychuk2021isaac} simulator demonstrate that our method achieves superior performance in Anymal and Humanoid tasks. Visualization results confirm that the robot behaviors learned through our method are consistent with those depicted in the input videos—a feat difficult to achieve with existing methods that use text as the learning target.

\end{itemize}

\section{Related work}
\subsection{Legged Robot Behavior Learning}
Legged robot behavior learning refers to the process of developing and optimizing a set of algorithms and techniques for a specified behavior to facilitate learning in legged robots. Research in this field can be categorized into two paradigms: one is imitation learning, which learns specific tasks by observing and mimicking the behaviors of humans or other intelligent entities. This often involves techniques such as behavior cloning~\cite{torabi2018behavioral, bratko1995behavioural} and inverse reinforcement learning~\cite{arora2021survey,hadfield2016cooperative}. Imitation learning is particularly crucial in the study of quadruped and humanoid robots, whose actions and tasks are often complex, making direct programming or teaching via traditional machine learning methods exceedingly challenging~\cite{kim2022humanconquad,PengCZLTL20,reske2021imitation}. For instance, Peng et al.~\cite{PengCZLTL20}focus on mimicking animal locomotion using meta-learning techniques in simulation environment to enhance the robot's adaptability in complex terrains. Reske et al.~\cite{reske2021imitation} employ model predictive control (MPC) to guide imitation learning, optimizing the control Hamiltonian to train robots in managing multiple gaits. Kim et al.~\cite{kim2022humanconquad} apply human motion data directly to robot control, integrating supervised learning with reinforcement learning to accommodate the robot’s dynamic properties. However, the success largely depends on the quality and applicability of the mimetic data.

The other paradigm is reinforcement learning (RL), which trains robots to learn specific tasks through a system of rewards and penalties. In the domain of quadruped and humanoid robots, RL is employed to enable robots to learn from their environment autonomously. Currently, RL has demonstrated potential~\cite{rudin2021cat,yang2020data,yue2020learning}. Yang et al.~\cite{yang2020data} enhance data efficiency by incorporating model-based RL and MPC, enabling robots to adapt to various tasks in real-world settings. Rudin et al.~\cite{rudin2021cat} focus on the issues of jumping and landing in low-gravity environments for legged robots. Nonetheless, a significant challenge in these RL approaches is designing effective reward functions. It relies on the expertise and intuition of domain specialists, a process that is not only time-consuming but also often fails to yield optimal solutions.

\subsection{Reward Designed by Large Language Model}
The advent of LLM has introduced new possibilities for automating the generation of reward functions~\cite{ma2023eureka, xie2023text2reward, yu2023language, KwonXBS23, lin2022inferring, hu2023language,Tang0TZFH23}. The concept of transforming natural language instructions into reward functions has been partially explored and advanced in prior research, as evidenced in studies such as text2reward~\cite{xie2023text2reward}. L2R~\cite{yu2023language} illustrates how reward functions can be derived from natural language task descriptions through LLM. Eureka~\cite{ma2023eureka} leverages the code of reinforcement learning environments as prompts for LLM, facilitating the automatic generation of task-specific reward functions. However, the task descriptions in existing works are overly generalized, failing to adequately constrain behavior characteristics. In contrast, our study proposes the generation of reward functions through the integration of video inputs. This innovation enables the reward function to more comprehensively capture the nuances of real dynamic environments and subtle behavioral changes, allowing for more precise adjustments in behavior to meet task requirements.

\begin{figure*}[!t]
    \centering
    \includegraphics[width=\textwidth]{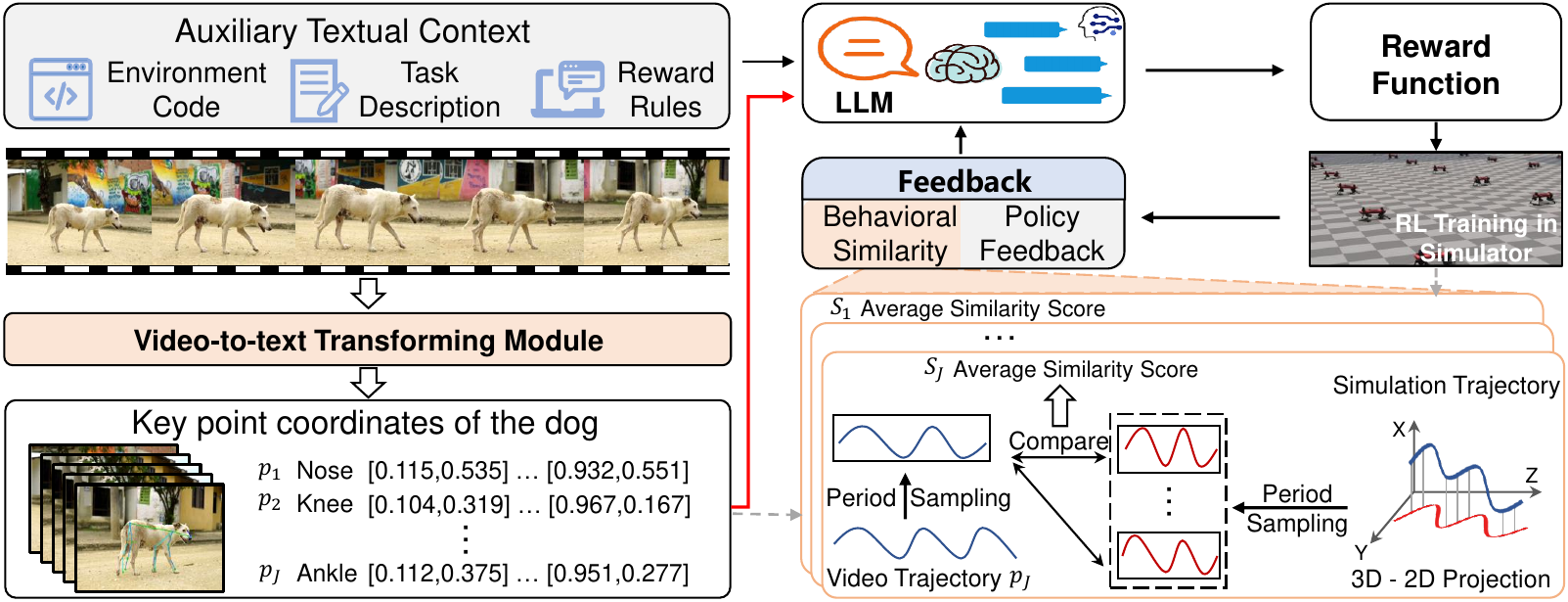}
    \vspace{-0.3cm}
    \caption{Overview of our Video2Reward method. We propose a video-to-text transforming module that converts keypoints of motion in video clips into text. This textual information, alongside auxiliary textual contexts, guides a large language model (LLM) in designing reward functions. After each training cycle, we visually compare the robot's keypoint trajectories with those of the target in the video, receiving feedback on the behavioral similarity. Lastly, LLM refines the reward functions based on this feedback and basic policy feedback, thereby directing the robot to learn specific behaviors demonstrated in the video.}
    \vspace{0.5cm}
    \label{fig2:method}
\end{figure*} 

\section{Preliminary}
\noindent \textbf{Behavior learning via reinforcement learning (RL).} 
Legged robot behavior control can be formulated as a Markov Decision Process (MDP) where the robot interacts with environment $E$. The MDP is represented as $(\mathcal{S},\mathcal{A},\mathcal{P}, R,\gamma)$ with state space $\mathcal{S}$, action space $\mathcal{A}$, transition probability function $\mathcal{P}$, reward function $R$, and discount factor $\gamma \in (0,1]$.
At each step $t$, the robot observes state $s_{t}$, takes action $a_t$ based on policy $\pi(s_t)$, transitions to next state $s_{t+1}$, receives reward $r_t = R(s_t,a_t,s_{t+1})$, and updates its state. The return from a state $s_t$ is defined as the accumulated $\gamma$-discounted reward: {$\sum\nolimits_{i = t}^T {{\gamma ^{i - t}} r_i}$}. The objective of RL is to optimize the policy $\pi$ by maximizing the expected return from the initial state. The design of reward functions is crucial in RL, relying on the experience of researchers and practitioners for manual trial-and-error refinement \cite{sutton2018reinforcement, ma2023eureka}. This is particularly challenging for tasks involving legged robot behavior learning with multiple degrees of freedom, where considerations such as robot stability, safety, energy efficiency, and behavioral accuracy need to be simultaneously addressed \cite{huang2023rewadaptive}.


\noindent \textbf{LLM-based reward function generation.}
To address the challenge of designing complex reward functions, large language model (LLM) can be utilized to generate executable reward function code for direct implementation within reinforcement learning frameworks. This approach, exemplified by methods such as L2R~\cite{yu2023language} and Eureka~\cite{ma2023eureka}, can be formulated as:
\begin{equation}\label{eq:eureka}
{R} = f_{LLM}(T_{ext}),
\end{equation}
where \( T_{ext} \) can correspond to a textual task description in L2R, or \(T_{ext} = [t_d, t_e, t_r]\) in Eureka, consisting of a task description \(t_d\), environment code \(t_e\), and coding rule \(t_r\). 
To further enhance the quality of the reward function, some studies also incorporate feedback loops. For instance, 
designing a feedback function \(F\) to evaluate the reward function generated in the \((n-1)^{th}\) iteration. The evaluation results are then converted back into text, re-fed into the LLM, and the process iterates to refine the reward function for the \(n^{th}\) round as
\begin{equation} \label{eq:eurekafeedback}
R^n = f_{LLM}(T,R^{n-1} ,F(R^ {n-1})).
\end{equation}



\section{Method}






\subsection{General Scheme of Our Approach}
Our focus is on addressing the challenges inherent in methods that use LLM to generate reward functions based on behavior learning goals provided in textual form. These methods often fail to accurately capture the detailed nuances of target behaviors, resulting in learning processes that are difficult to control and inefficient for acquiring specific actions. Our principal idea is to incorporate video inputs depicting the desired behaviors into the reward generation process. By utilizing motion information from the videos—information that is often challenging to articulate using language alone—we aim to provide more concrete, physically plausible learning objectives. This, in turn, facilitates the creation of a reward function better suited for legged robot behavior learning, as depicted in Figure \ref{fig2:method}.

Formally, given a video \(v\) that includes the target behavior, the first step involves inputting this video into a video-to-text module (\(M_{v2t}\)). This module converts the motion information of the target into the coordinates of the moving object's joints and presents the trajectories of each joint’s coordinates as input to the LLM. This is represented as a transformation of Equation (\ref{eq:eureka}) into
\begin{equation}\label{equation:3}
R = f_{LLM}(T_{ext}, M_{v2t}(v)).
\end{equation}
We apply the executable reward function code $R$ within a reinforcement learning environment (such as IsaacGym~\cite{makoviychuk2021isaac}) to train a policy for behavior learning. Subsequently, leveraging the in-context improvement capability of LLM, we design a mechanism for multi-round iterative updates of the reward function, which acquires feedback from the trained policy to enhance $R$. Benefiting from our approach of depicting the learning target through specific video descriptions—unlike traditional methods that derive numerical indices directly from policy performance—we propose a novel video-assisted feedback function $F_{fb}$. This function evaluates the behavioral mimicry visually, generating textual feedback and thereby evolving the iterative nature of the reward function:
\begin{equation}
R^n = f_{LLM}(T, R^{n-1}, M_{v2t}(v), F_{fb}(R^{n-1}, \pi^{n-1}, v)).
\end{equation}

In the following, we aim to answer two questions: 1) how to generate reward with video input? (Section~\ref{Sec:v2t}), and 2) how to iteratively refine reward w.r.t the learning target in the video? (Section~\ref{Sec:refine}).

\subsection{Video2Text for Generating Rewards via LLM}\label{Sec:v2t}
To enable LLM to automatically generate rewards based on objects described in videos, the key lies in the conversion of visual motion information into textual data that can be interpreted by LLM. 

\noindent\textbf{Video-to-text transforming module $M_{v2t}$.}
To allow LLM to focus more effectively on the patterns and rules inherent to the motion itself, we use a sequence of object joint trajectories to represent motion information. The underlying rationale includes several aspects: \textbf{First}, joint trajectories capture the essential characteristics of motion while omitting unnecessary background details, providing a precise description of the motion state and changes of the target object, such as position and velocity. This provides robots with exact motion targets and references, aiding in more precise motion control and simulation. \textbf{Second}, joint trajectories can be expressed in coordinate form, offering an easily interpretable and processable data format for LLM. \textbf{Third}, using joint trajectories as input standardizes the data across different video sources or motion tasks into a uniform format.

Given a video, we uniformly sample the video to obtain a set of frames \(\mathcal{I} = \{I_l\}_{l=1}^L\). For each frame \(I_l\), keypoints can be extracted using a pose estimation method, such as RTMpose~\cite{cai2020learning} for human and HRNet~\cite{sun2019deep} for animal, resulting in a set of \(J\) keypoints for the target object. Let $p_j(l) = (x_{lj}, y_{lj})$ be the coordinates of the $j$-th keypoint in $I_l$, then the trajectory sequence of the $j$-th keypoint can be represented as 
$\mathcal{T}_j = \{ p_j(l) \mid l \in \{1, 2, \ldots, L\} \}$. In this way, $\{\mathcal{T}_j\}_{j=1}^J$ can be regarded as a motion descriptor for the target behavior.

\noindent\textbf{Auxiliary Textual Context $\bm{T_{aux}}$.}
In addition to visual inputs, it is essential to incorporate auxiliary textual context when interacting with LLM. This context includes the following components:

\textbf{1) Environment Code $\bm{t_e}$:} The raw code of the reinforcement learning environment encompasses essential details that define the state \( S \), action \( A \), and transition logic. These elements are crucial for the generation of the reward function \( R \). By leveraging LLM' capabilities in understanding code structure and syntax, key task variables and potential reward signals are identified, enabling the automatic generation of efficient and precise task-specific reward functions. For instance, if the environmental simulation pertains to a humanoid robot scenario, the environmental code would include variables such as the position $p$ and velocity $v$ of the robot's limbs. LLM uses this information to generate a reward function, aimed at optimizing specific actions of the humanoid robot, such as enhancing running efficiency or increasing the smoothness of movements.


\textbf{2) Task description $\bm{t_d}$:} A brief description of the tasks we are undertaking informs the LLM of our goals for each task.
For example, in a quadruped robot behavior learning task, the reward function must ``make the quadruped follow randomly chosen $x$, $y$, and yaw target velocities''. In bipedal robots, the reward function should ``make the humanoid run as fast as possible''.

\textbf{3) Generation rule $\bm{t_r}$:} Following the Eureka framework, we define a set of explicit reward design rules, such as: ``reward function should use useful variables from the environment as inputs.'' These rules direct LLM to effectively utilize environmental variables and ensure compatibility with TorchScript. This increases the feasibility and practicality of generated code, enabling smooth deployment and training within the reinforcement learning environment.  

Lastly, we concatenate the textual motion information obtained from the video-to-text transforming module with the auxiliary textual context to form a prompt inputted into the LLM, requesting it to generate an executable reward function.

\subsection{Video-Assisted Iterative Reward Refinement}\label{Sec:refine}

Despite the in-context comprehension and generation capabilities of LLM, ensuring quality in single-round reward generation remains challenging. In practice, the expected outcomes of LLM often result from its multiple interactions with the user, as LLM can modify its responses based on user feedback. Inspired by this, we propose an iterative reward refinement process, which includes evaluating the effectiveness of a generated reward function and providing feedback to the LLM, serving as a reference for adjustments and enhancing the quality of the reward function.

To obtain feedback, one may record rewards during the training process or assess the agent’s performance on its textual task description. For example, in the Anymal task, the difference between the robot's linear velocities and the target velocities serves as feedback in Eureka~\cite{ma2023eureka}. However, textual task descriptions can be overly generalized for specific behaviors; for instance, in Humanoid tasks, while the agent might excel in ``running fast'', its running posture may not accurately reflect human running postures.

\noindent \textbf{Video-assisted reward function evaluation.}
Fortunately, thanks to video inputs, we can directly measure the differences between the target behaviors and those learned by the agent, obtaining more accurate feedback. Consequently, we introduce a video-assisted feedback function, which correlates the trained agent’s movements with the target behaviors $\mathcal{T}$ observed in videos.

Specifically, we \textbf{first} test the trained policy in a simulated environment, continuously collecting the 
trajectory of all $J$ keypoints over $T$ steps as $\{\hat{\mathcal{T}}_j\}_{j=1}^J$, where $\hat{\mathcal{T}}_j = \{ \hat{p}_j(t)=(x_{tj}, y_{tj}, z_{tj}) \}_{t=1}^T$. 
Since a direct comparison between 3D trajectory $\hat{\mathcal{T}}_j$ and 2D trajectory $\mathcal{T}_j$ collected from the input video is not feasible, we project the 3D coordinates of $\hat{\mathcal{T}}_j$ along the direction of motion, converting them into 2D coordinates. \textbf{Second}, to enable a more precise comparison between two trajectories, we use the autocorrelation function~\cite {breitenbach2023method} to determine the trajectory’s periodic length, segmenting the trajectory into multiple segments, each of which contains a trajectory with two periods. We then compare the segmented trajectory via a similarity calculation function $F_{sim}$. The similarity scores across segments are then averaged to evaluate the overall similarity between $\hat{\mathcal{T}}_j$ and $\mathcal{T}_j$. Here, we adopt the FastDTW algorithm~\cite{salvador2007toward} as $F_{sim}$, which efficiently handles time series with variable period lengths. To achieve a finer granularity of fitness values, the trajectory similarity for each joint is calculated individually, \( S_j = F_{sim}(\hat{\mathcal{T}}_j, \mathcal{T}_j)\), thus each agent ultimately receives \( J \) similarity score. In testing, multiple agents are evaluated, and their results are averaged, providing feedback in the form of a similarity score list that can be inputted into LLM.

\subsection{Training Details}
We conduct \(N\) rounds of feedback iteration. In each round, we require LLM to produce a set of \( K \) reward functions \(\{r_1, r_2, \ldots, r_K\} \). This strategy helps to reduce the likelihood that all reward functions are simultaneously erroneous. Subsequently, each reward function is used for training via the proximal policy optimization (PPO) strategy~\cite{Schulman_Wolski_Dhariwal_Radford_Klimov_2017}, resulting in policies \(\pi_k = \textit{PPO}(r_k)\). Throughout this training phase, we evaluate the performance of trained policies using the max training success metric $H_{mts}$ defined in Section~\ref{Metric}, represented as  \(s_k = H_{mts}(\pi_k)\). We select reward \(r_{\text{best}}\) and policy \(\pi_{\text{best}}\) that yield the highest \(s\) and apply \(\pi_{\text{best}}\) to sample robot trajectories $\hat{\mathcal{T}}$. These trajectories are then compared with video trajectories using a similarity calculation \(D = F_{\text{sim}}(\hat{\mathcal{T}},\mathcal{T}\)). Finally, \(r_{\text{best}}\), \(s_{\text{best}}\), \(D\), and the auxiliary textual context $T_{aux}$ are combined and returned to LLM to generate the next round's reward functions. After \(N\) rounds of iterations, the optimal reward function \(r^N\) and policy \(\pi^N\) from the entire iterative refinement process are outputted, as detailed in Algorithm~\ref{algorithm:V2R}.


\begin{algorithm}[!t] 
\caption{Our Proposed Video2Reward Method} 
\begin{algorithmic}[1]
\label{algorithm:V2R}
\REQUIRE{Motion trajectories from video $\mathcal{T}=M_{v2t}(v)$, Large language model $f_{LLM}$, Auxiliary textual context $T_{aux}$}
\STATE \textbf{Hyperparameters}: search iteration $N$, reward samples $K$
\FOR{$N$ iterations}
    \STATE \textcolor{darkgreen}{\footnotesize\texttt{// Generate $K$ reward functions from LLM}}
    \STATE $R_1^n, R_2^n, \ldots, R_{K}^n \sim f_{LLM}(T_{aux}, \mathcal{T})$
    \STATE \textcolor{darkgreen}{\footnotesize\texttt{// Train RL policies with PPO}}
    \STATE $\pi_1^n = PPO(R_1^n),\ldots,\pi_K^n = PPO(R_K^n)$
    \STATE \textcolor{darkgreen}{\footnotesize\texttt{// Evaluate and select the best policy}}
    \STATE $best = \mathop{\arg\max}_{k} \{s_1,\ldots,s_K\}$, where $s_k= H_{mts}(\pi_{k}^n)$
    \STATE \textcolor{darkgreen}{\footnotesize\texttt{// Sample robot trajectories from simulator}}
    \STATE $\hat{\mathcal{T}} \sim \pi_{best}^n$
    \STATE \textcolor{darkgreen}{\footnotesize\texttt{// Calculate the similarity between $\mathcal{T}$ and $\hat{\mathcal{T}}$}}
    \STATE $d= F_{sim}(\hat{\mathcal{T}},\mathcal{T})$
    \STATE \textcolor{darkgreen}{\footnotesize\texttt{// Update the input of LLM}}
    \STATE $T_{aux} := T_{aux} + R_{best}^{n} + s_{best}^{n} + d$
    \STATE \textcolor{darkgreen}{\footnotesize\texttt{// Update the best reward function}}
    \STATE $R^{n}, s^{n} = \begin{cases} 
  (R^n_{best}, s^n_{best}), & \text{if } s^n > s^{n-1}_{best} \\
  (R^{n-1}, s^{n-1}), & \text{otherwise}
\end{cases}$
    \ENDFOR
\STATE \textbf{Output:} $R^N$
\end{algorithmic}
\end{algorithm}

\section{Experiments}

\subsection{Environments}
For fair comparisons with other methods, we use IsaacGym~\cite{makoviychuk2021isaac}, a high-performance simulation environment developed by NVIDIA for robotics research and development. This platform includes a series of predefined environments that encompass scenarios involving robotic arms, quadruped robots, bipedal robots, and more. In our research, we evaluate our methods using the quadrupedal (Anymal task) and bipedal (Humanoid task) robots.

\subsection{Baseline methods}
\noindent\textbf{Eureka~\cite{ma2023eureka}.}
It uses language to specify targets, incorporating environmental code as context and employing LLM to generate reward functions. For fair comparisons, we maintain the same parameter settings as Eureka in our experiments. The training process is identical to that of Eureka, consisting of 5 iterations, during each of which 16 reward samples are generated, using its open-source code.

\noindent\textbf{Human~\cite{makoviychuk2021isaac} .} 
 In the IsaacGym simulator we employ, each task incorporates a reward function designed by expert-level researchers in reinforcement learning. For more details, please refer to~\cite{makoviychuk2021isaac}.

\noindent\textbf{Sparse~\cite{ma2023eureka}.} Following Eureka, we employ the sparse reward method as a baseline. While this approach involves a shaped reward, it is less refined compared to the human-derived reward.
In the Anymal task, the sparse reward is defined as $R_{S\_Anymal} = -(E_{lin\_vel} + E_{ang\_vel})$. Here, $E_{lin\_vel}$ represents the discrepancy between the robot's actual linear velocity and its target linear velocity, while $E_{ang\_vel}$ indicates the error between the robot's actual angular velocity and its target angular velocity. In the Humanoid task, the sparse reward is denoted as $R_{S\_Humanoid} = R_{p}$, where $R_{p}$ quantifies the incentive provided to the robot to move towards a target location within the environment.

\subsection{Evaluation Metric} \label{Metric}

\noindent \textbf{Maximum training success $H_{mts}$.} The term $H_{mts}$ represents the average of sparse reward and conveys different meanings in the Anymal and Humanoid tasks. In the Anymal task, $H_{mts}^{a}$ is a reward metric that quantifies the error in velocity control. A higher value of $H_{mts}^{a}$ indicates greater precision and stability in the robot's motion control. In the Humanoid task, $H_{mts}^{h}$ is used to measure the Humanoid robot's movement toward a target direction. A higher $H_{mts}^{h}$ means the robot moves faster toward the target direction. Across all tasks, a higher $H_{mts}$ denotes that the robot's performance more closely aligns with the task requirements.

\noindent \textbf{Human normalized score.} Like Eureka, we employ the Human normalized score to evaluate the overall performance of robots on tasks:
\begin{equation} \label{eq:humanns}
\frac{\text{Method} - \text{Sparse}}{\left| \text{Human} - \text{Sparse} \right|}.
\end{equation}
In this formula, ``Method'' refers to the approach being evaluated. ``Method'', ``Sparse'', and ``Human'' each represent the values of $H_{mts}$ obtained through training with three different methods.

\noindent \textbf{DTW score.} The Dynamic Time Warping (DTW) score is employed as a metric to measure the similarity between behavioral trajectories. A higher DTW score indicates a lower similarity between the robot's behavior and that of the target in the video; conversely, a lower DTW score implies more similarity.

\subsection{Implementation Details}
\noindent\textbf{Legged robot's task description.} 
In assigning task descriptions for both the legged robot models, Anymal and Humanoid, we follow the same criteria as those employed in Eureka. For Anymal, the objective is to have the quadruped follow randomly chosen $x$, $y$, and yaw target velocities, while for the Humanoid, the aim is to maximize its running speed. In selecting behaviors, we focus on two distinct ones—amble and running—for Anymal, given the significant differences in movement postures and behavioral characteristics between these modes. For the Humanoid task, running serves as the sole behavior for robot learning. Throughout subsequent experiments within the Anymal project, unless explicitly stated otherwise, the experimental metrics are calculated as the average of the two behaviors.


\begin{figure*}[t] 
    \centering 
    \includegraphics[width=.3\linewidth]{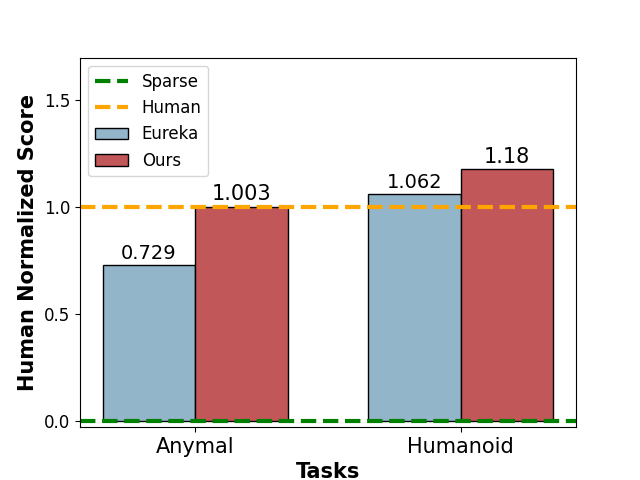} 
    \includegraphics[width=.3\linewidth]{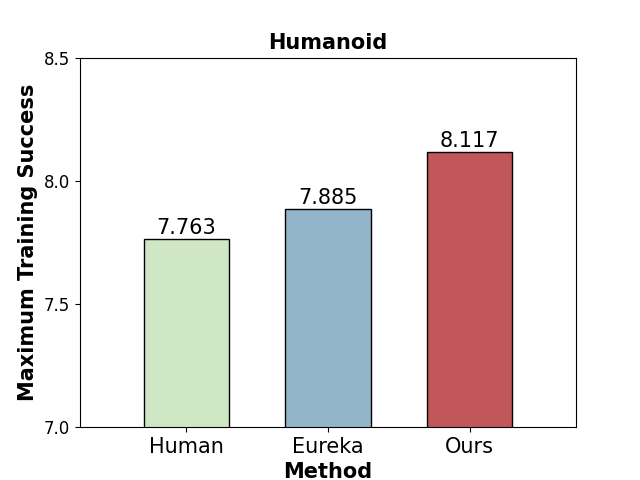} 
    \includegraphics[width=.3\linewidth]{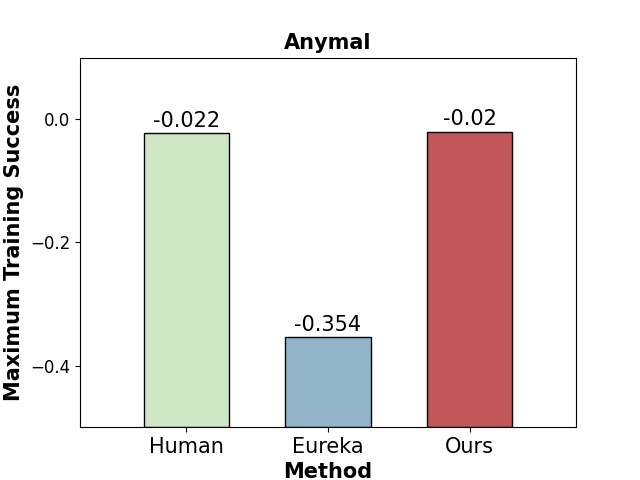} 
    \vspace{0.1cm}
    \caption{Comparisons on Anymal and Humanoid tasks in terms of human normalized score and maximum training success. Our method outperforms all baseline methods, including Human, suggesting the introduction of visual information is effective.}
    \vspace{0.3cm}
    \label{fig:bench} 
\end{figure*}

\noindent\textbf{Video processing details.}
We selected video clips from the internet that cover 2-3 complete motion cycles of the target, in order to capture the continuity of the actions. For the Humanoid task, we chose to capture the coordinates of the object's joints every three frames, obtaining 2 seconds of video information. This approach ensures that the collected coordinate data includes 2 to 3 running cycles of a human. In the Anymal task, our video processing method involves capturing the coordinates of the object's joints every nine frames, totaling 3 seconds of video information. For videos shot with a stationary camera, we employ a normalization method based on the dimensions of the video frame to preserve the motion trajectory maximally. For videos recorded with a non-stationary camera, we use a normalization method based on the object's bounding box, as it enables the capture of the target's motion information while minimizing interference from the moving camera. Please refer to the supplementary material for more details\footnote{Supplementary material: \url{https://github.com/Alvin-Zeng/Video2Reward}}.

\begin{figure}[t] 
    \centering 
    \includegraphics[width=.98\linewidth]{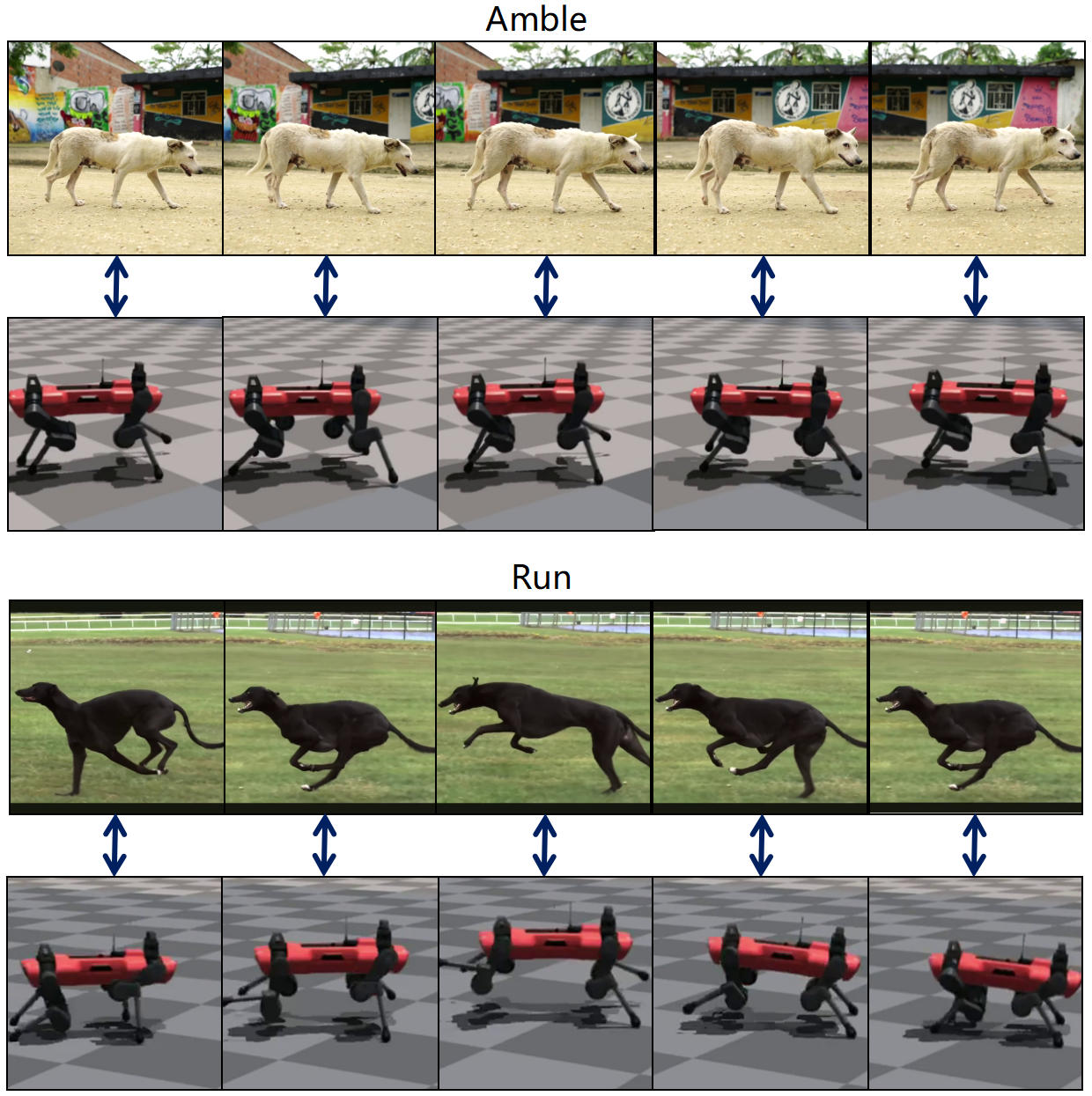} 
    \vspace{0.1cm}
    \caption{Visualization results of quadrupedal robots trained by our method. Specific behaviors are learned effectively from different videos.} 
    \vspace{0.6cm}
    \label{fig:anymal_behavior} 
\end{figure}

\subsection{Experimental Results}
\noindent\textbf{Comparisons with state-of-the-art LLM-based methods.} 
We compared our method with other LLM-based reward generation methods on two typical legged robotic behavior learning tasks involving the Anymal (a quadruped robot) and Humanoid (a bipedal humanoid robot). To ensure fair comparisons with Eureka, we replicated their method using the official code, maintaining consistent experimental settings and parameters. As demonstrated in Figure \ref{fig:bench}, the models trained using our approach exhibited superior performance across all metrics, surpassing both Eureka and expert-designed baselines. Notably, our method achieves a 37.6\% improvement over Eureka on the Anymal task and an 11.1\% enhancement on the Humanoid task in terms of the average human-normalized score.

\begin{table}[t]
    \caption{Comparisons w.r.t the behavioral similarity between the learned behavior and the target behavior, measured by DTW score. \textbf{Lower values represent the greater similarity.}}
    \vspace{0.6cm}
\centering
\label{tab:Behavioral Similarity Evaluation}
\resizebox{.6\columnwidth}{!}{
\centering 
\begin{tabular}{lccc} 
\toprule
\multirow{2}{*}{Method} & \multicolumn{2}{c}{Anymal} & \multicolumn{1}{c}{Humanoid} \\
\cmidrule(r){2-3} \cmidrule(r){4-4}
                        & Amble $\downarrow$ & Run $\downarrow$ & Run $\downarrow$ \\
\midrule
Human~\cite{makoviychuk2021isaac}         & 26.368        & 64.117        & 8.721 \\ 
Eureka~\cite{ma2023eureka}       & 34.186        &  94.542       & 8.237  \\ 
Eureka-$t_d$ & 44.727        & 130.427      & 8.252   \\ 
\textbf{Ours}         & \textbf{17.292}        & \textbf{28.124}        & \textbf{7.359}  \\ 
\bottomrule
\end{tabular}
}
\end{table}

\begin{table}[t]
\centering
\caption {Evaluation of the trained policy on the reward function designed by experts. Our method performs better on metrics of interest to experts.}
\vspace{0.6cm}
\label{tab:Sub-Reward Comparison}
\resizebox{.65\columnwidth}{!}{
\begin{tabular}{lcc|cc}
\toprule
\multirow{2}{*}{Method}& \multicolumn{2}{c}{Anymal} & \multicolumn{2}{c}{Humanoid} \\
\cmidrule(r){2-3} \cmidrule(r){4-5}
                & \makecell{$r_{\text{lin}}$} & \makecell{$r_{\text{ang}}$} & 
                \makecell{$r_{up}$} & \makecell{$r_{alive}$} \\
\midrule
Eureka~\cite{ma2023eureka}              & 22.629    &  13.441  & 34.714    &  1726.317    \\
\textbf{Ours}    & \textbf{46.372}    & \textbf{22.599}   & \textbf{45.366}     &  \textbf{1789.592}    \\  
\bottomrule
\end{tabular}
}
\end{table}

\begin{table}[t]
\centering
\caption {Ablation study of video-assisted feedback (VAF).}
\label{tab:Visual Ablation}
\vspace{0.6cm}
\resizebox{\columnwidth}{!}{
\begin{tabular}{lcc|cc}
\toprule
\multirow{2}{*}{Method}& \multicolumn{2}{c}{Anymal} & \multicolumn{2}{c}{Humanoid} \\
\cmidrule(r){2-3} \cmidrule(r){4-5}
                & \makecell{Human \\normalized score} & \makecell{DTW score $\downarrow$} & 
                \makecell{Human \\normalized score} & \makecell{DTW score $\downarrow$} \\
\midrule
Eureka~\cite{ma2023eureka}              & 0.729 & 34.186   &1.062  & 94.542    \\
Ours w/o VAF        & 0.921 & 20.355   &1.165   & 60.083    \\
\textbf{Ours}                & \textbf{1.003} & \textbf{17.292}  &\textbf{1.180}  & \textbf{28.124}    \\
\bottomrule
\end{tabular}
}
\end{table}

\noindent\textbf{Q1: Is our learned behavior closer to target behavior?} To validate our method's ability to enable legged robots to learn specific behaviors from videos, we compare the similarity of the behaviors learned through various methods with the target behaviors demonstrated in the original videos. Specifically, for each keypoint of the robot, we calculate the dynamic time warping (DTW) score between its trajectory and the trajectory of the corresponding keypoint detected from the video. We then average the DTW score across all keypoints for comparison. Lower values represent the greater similarity.  From Table \ref{tab:Behavioral Similarity Evaluation}, the robot trained using our method across all behaviors in two types of tasks has significantly lower DTW scores than those obtained using the Eureka and Human. This result confirms that our approach effectively assists robots in learning specific behaviors from natural videos.

We also introduce a variant by modifying the task description in the Eureka method, termed Eureka-$t_d$, which is structured as: ``To make the \{quadruped/humanoid\} \{amble/run\} like real \{dogs/humans\}". From Table \ref{tab:Behavioral Similarity Evaluation}, the DTW score obtained via the Eureka-$t_d$ method is higher than those of Eureka itself. This is because the original task description for Eureka includes reference standards. For instance, in the Anymal task, robots are required to move at a random speed in the $x$ and $y$ directions, which provides clear guidance for LLM. However, when instructions are converted into more abstract linguistic descriptions, LLM fails to comprehend the mechanics of how a dog walks. This limitation arises because LLM is trained using textual data, rendering them incapable of generating suitable reward functions. Our method provides LLM with clear and detailed learning targets, facilitating the creation of effective reward functions that steer robots toward acquiring specific behaviors.

\noindent\textbf{Q2: Why does our method lead to improvement in behavior learning?}
To understand the enhancements that our method brings to behavior learning, we evaluate robots trained with our method and Eureka using reward functions designed by human experts. In the Anymal task, we evaluate rewards \(r_{lin}\) and \(r_{ang}\), which focus on linear and angular velocity errors, guiding the robotic dog to mimic specified velocities. For the Humanoid task, we employ \(r_{up}\) to encourage an upright posture for better motion control, and \(r_{alive}\) to indicate survival. These expert-crafted rewards encompass motion speed, posture, and other relevant factors, serving as benchmarks for evaluating robot behavior. Our method consistently outperforms Eureka, as shown in Table \ref{tab:Sub-Reward Comparison}, with trained robots earning significantly more rewards across various motion metrics. Specifically, in the Anymal task, our method yields over double the linear speed reward compared to Eureka, while also significantly improving angular velocity and the Humanoid's Up reward. These findings affirm our method's superiority in metrics important to experts.
\begin{figure}[t] 
    \centering \includegraphics[width=.8\linewidth]{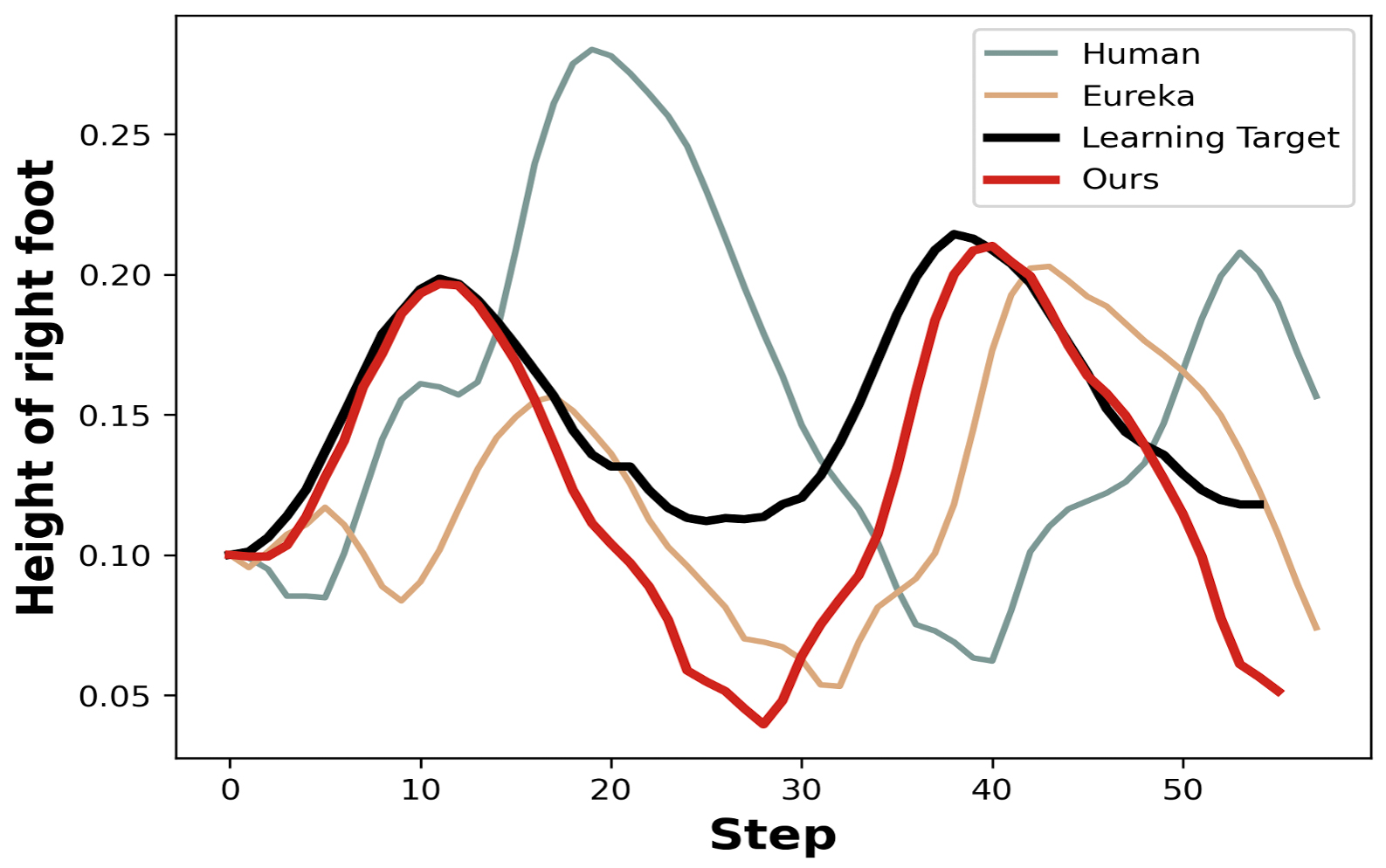} 
    \vspace{0.1cm}
    \caption{Comparisons of the right foot trajectories of Humanoid robots.} 
    \vspace{0.6cm}
    \label{fig:curve} 
\end{figure}

\noindent \textbf{Q3: Do we need video-assisted feedback?}
 A significant distinction between our method and existing approaches lies in the introduction of video as an input and the utilization of this visual information to form feedback. To investigate the efficacy of this feedback, we designed one variant, \textbf{Ours w/o video-assisted feedback (VAF)}: video is included as input, but is not considered when computing feedback. As shown in Table \ref{tab:Visual Ablation}, for the Anymal task, our original method achieves a Human normalized score of 1.003. When we remove the visual information feedback, the score decreases to 0.921. The performance of this variant still surpasses that of Eureka, and the best performance is achieved when video is considered both as an input and in feedback, exceeding the rewards designed by human experts. This underscores the advantage of visually assessing the performance of trained policies to enhance the quality of reward generation.
\subsection{Qualitative Results}
\noindent \textbf{Keypoint trajectory.} To validate whether our method facilitates the generation of smoother and more natural movements in robots, we focus specifically on the motion trajectories of the robots' feet in the Humanoid task. We compare these trajectories with those from video inputs through visualizations. This choice is predicated on the rationale that the trajectory of the feet reflects the robot's gait and posture. As illustrated in Figure \ref{fig:curve}, the feet trajectories of robots trained with our method are noticeably smoother and more closely resemble the video trajectories compared to those trained with Eureka and human rewards. This demonstrates its potential to enhance the smoothness of robot motions and their alignment with natural movement patterns.

\noindent \textbf{Performance on quadrupedal robots behavior learning.}
To verify the effectiveness of our method in enabling robots to learn specific behaviors from videos, we observe and visually display the actual behaviors of robots within a simulation environment. Figure \ref{fig:anymal_behavior} demonstrates that our approach effectively enables quadruped robots to adopt behaviors such as "amble" and "run." Notably, the robots exhibited the continuous alternating footfalls characteristic of ambling, alongside the propulsion of hind legs and the elevation and extension of front legs observed in running.

\begin{figure}[t] 
    \centering 
    \includegraphics[width=.98\linewidth]{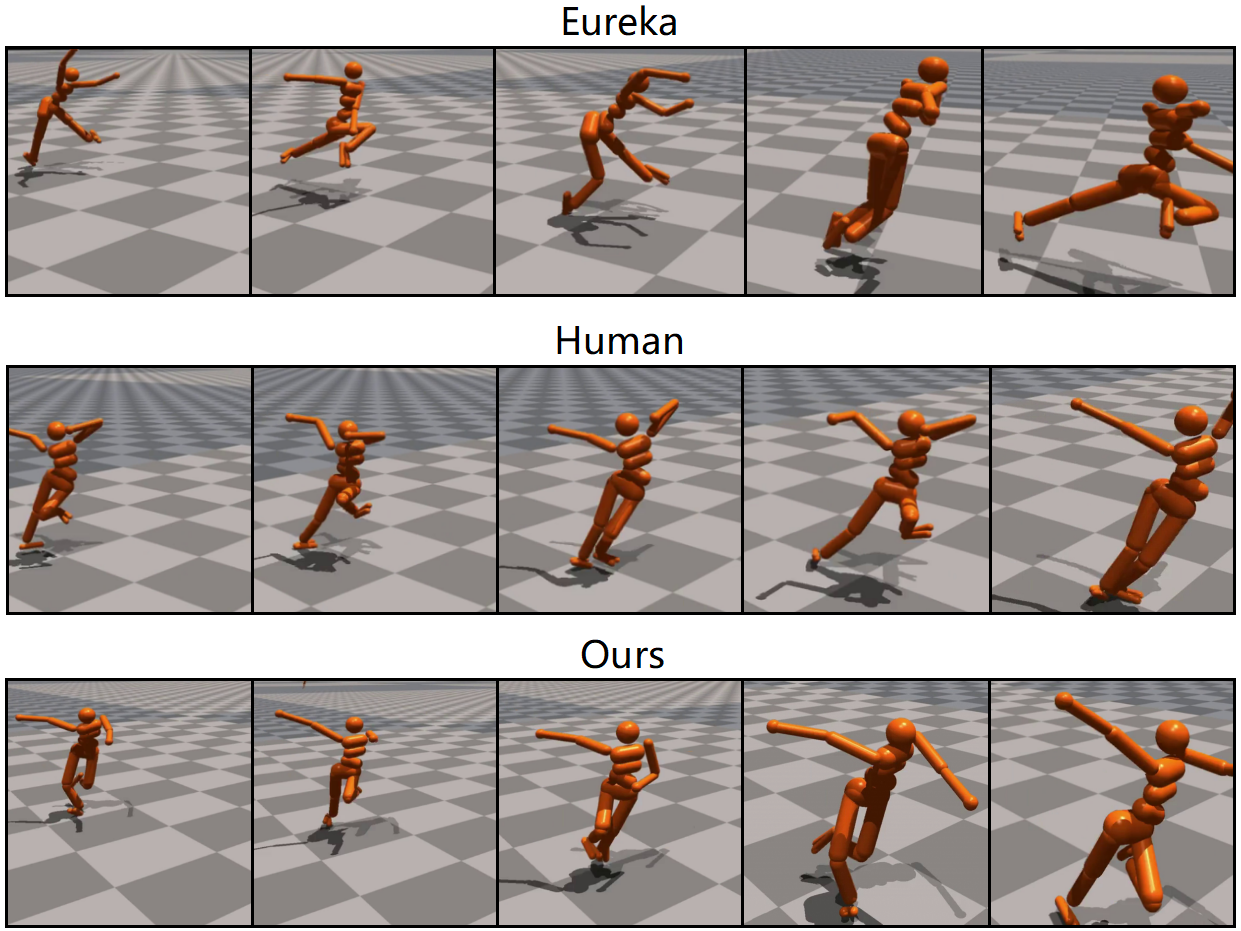} 
    \vspace{0.1cm}
    \caption{Comparisons of the learned running behavior in the Humanoid task. Our running pose is visually the closest to that of a real human.} 
    \vspace{0.6cm}
    \label{fig:humanoid_behavior} 
\end{figure}

\noindent \textbf{Performance on humanoid robots behavior learning.} Figure \ref{fig:humanoid_behavior} displays the running postures of humanoid robots trained using three different methods. Notably, the robots trained with our method exhibit a gait and arm swing more closely resembling human running behaviors. More details can be found in the supplementary demonstration video\footnotemark[\value{footnote}].

\section{Conclusions}

In conclusion, our study addressed the challenge of precise behavior learning in legged robots by proposing a new video2reward approach that harnesses the capabilities of the large language model (LLM) to generate highly accurate reward functions directly from video inputs. Unlike existing LLM-based reward generation methods that take only text as input, we leveraged real-world motion data to provide a more detailed and dynamically adaptable learning framework. By extracting keypoint trajectories from videos of desired behaviors and using these to inform the LLM-generated reward functions, our system ensured that the learned behaviors closely mimic the target movements. This methodology facilitated the rapid adaptation to diverse behaviors such as walking and running, which are critical for legged robots. The effectiveness of our method is validated through rigorous testing in bipedal and quadrupedal robot motion control tasks, where it has demonstrated superior performance, exceeding previous best LLM-based reward generation methods by more than 37.6\% according to human normalized scores. This marked a step forward in the field of robotics, promising advancements in how robots learn and adapt to complex tasks in real-world scenarios.




\begin{ack}
This work was supported by the National Natural Science Foundation of China (NSFC) under Grants 62202311; The Shenzhen Natural Science Foundation (the Stable Support Plan Program) under Grant 20220809180405001 and 20231122104038002; Excellent Science and Technology Creative Talent Training Program of Shenzhen Municipality under Grant RCBS20221008093224017; The Guangdong Basic and Applied Basic Research Foundation under Grants 2023A1515011512.
\end{ack}



\bibliography{mybibfile}

\end{document}